\documentclass[11pt]{article}

\usepackage[final]{acl}
\usepackage{multirow}
\usepackage{float}
\usepackage{times}
\usepackage{latexsym}
\usepackage{array}
\usepackage{amsmath} 
\usepackage[T1]{fontenc}
\usepackage{amssymb}
\usepackage{booktabs}
\usepackage[utf8]{inputenc}

\usepackage{microtype}

\usepackage{inconsolata}

\usepackage{graphicx}

\usepackage{dblfloatfix}


\title{Consistency-Aware Parameter-Preserving Knowledge \\ Editing Framework for Multi-Hop Question Answering}

\author{
  \textbf{Lingwen Deng\textsuperscript{1}},
  \textbf{Yifei Han\textsuperscript{2}},
  \textbf{Shijie Li\textsuperscript{2}},
  \textbf{Yue Du\textsuperscript{2}},
  \textbf{Bin Li\textsuperscript{2,*}}
\\
\\
  \textsuperscript{1}Vanderbilt University \\
  \textsuperscript{2}Shenzhen Institute of Advanced Technology, Chinese Academy of Sciences \\
  \small{\textsuperscript{*}Corresponding author}
}

\begin{document}
\maketitle
\begin{abstract}
 Parameter-Preserving Knowledge Editing (PPKE) enables updating models with new information without retraining or parameter adjustment. Recent PPKE approaches used knowledge graphs (KG) to extend knowledge editing (KE) capabilities to multi-hop question answering (MHQA). However, these methods often lack consistency, leading to knowledge contamination, unstable updates, and retrieval behaviors that are misaligned with the intended edits. Such inconsistencies undermine the reliability of PPKE in multi-hop reasoning. We present CAPE-KG, Consistency-Aware Parameter-Preserving Editing with Knowledge Graphs, a novel consistency-aware framework for PPKE on MHQA. CAPE-KG ensures KG construction, update, and retrieval are always aligned with the requirements of the MHQA task, maintaining coherent reasoning over both unedited and edited knowledge. Extensive experiments on the MQuAKE benchmark show accuracy improvements in PPKE performance for MHQA, demonstrating the effectiveness of addressing consistency in PPKE. All the codes are available at \href{https://anonymous.4open.science/r/CAPE-KG}{this link}.
\end{abstract}

\section{Introduction}
Knowledge editing \citep{wang2023survey_kme} enables large language models (LLMs) to incorporate new or corrected information without the cost of full retraining. KE approaches can be divided into parameter-updating methods and parameter-preserving methods \citep{li2024learning}. Parameter-updating methods, such as ROME \citep{meng2022locating}, MEMIT \citep{meng2023mass}, and T-Patcher \citep{huang2023transformerpatcher}, directly modify model weights to encode edits, which are computationally expensive. In contrast, PPKE avoids changing parameters and instead relies on external structures or contexts, offering greater transparency and stability. IKE \citep{zheng2023ike} edits knowledge through in-context demonstrations, MeLLo \citep{meng2023mass} supports multi-hop editing through query decomposition and fact retrieval, and PokeMQA \citep{gu2024pokemqa} performs decomposition and conflict checking. PPKE offers a cheaper and more efficient paradigm by avoiding both retraining and parameter modification. While early efforts in PPKE primarily addressed simple single-hop factual updates \citep{zheng2023ike}\citep{zhang2024comprehensive}, research has extended its scope to multi-hop question answering (MHQA) \citep{mavi2022survey}\citep{cheng2024temporal}, a more challenging setting that requires models to integrate edited knowledge with existing facts through multi-step reasoning. Recent work \citep{lu2025kedkg} adopts KG-based PPKE by injecting edited knowledge into a knowledge graph and retrieving relational paths, enabling models to perform explicit reasoning under edits for MHQA.

However, current KG-based PPKE is facing a critical weakness: lacking consistency \citep{cohen2024ripple}. Specifically, it struggles to maintain a knowledge boundary between factual and edited knowledge, to guarantee stable and conflict-free updates, and to ensure retrieval remains aligned with the intent of edit \citep{zhao2024ripplecot}. Inconsistency across these dimensions leads to less effective and reliable KE \citep{zhang2024locate}. In this paper, we propose CAPE-KG, Consistency-Aware Parameter-Preserving Editing with Knowledge Graphs, the first consistency-aware framework for PPKE. It improves consistency across the entire pipeline, ensuring that KG construction, knowledge update, and knowledge retrieval are all aligned with the PPKE objective. Our main contributions are:
\vspace{-0.2cm}
% We introduce a multi-layer KG architecture ensuring the consistent knowledge boundary between factual and edited knowledge during PPKE.

% We design case-isolated updating mechanisms with conflict arbitration that guarantee edits are applied atomically and align with the case-isolated assumption.

% We present an edit-aware retrieval module with progressive strategies and edit-related retrieval routing, ensuring the retrieval process is always consistent with the intended KE.

% Extensive experiments on the MQuAKE benchmark\citep{zhong2023mquake} demonstrate that CAPE-KG consistently outperforms existing baselines, achieving substantial gains in consistency and accuracy across backbones.

\begin{enumerate}
    \item We introduce a multi-layer KG architecture ensuring the consistent knowledge boundary between factual and edited knowledge during PPKE.
    \vspace{-0.2cm}
    
    \item We design case-isolated updating mechanisms with conflict arbitration that guarantee edits are applied atomically and align with the case-isolated assumption.
    \vspace{-0.2cm}
    
    \item We present an edit-aware retrieval module with progressive strategies and edit-related retrieval routing, ensuring the retrieval process is always consistent with the intended KE.
    \vspace{-0.2cm}
    
    \item Extensive experiments on the MQuAKE benchmark~\citep{zhong2023mquake} demonstrate that CAPE-KG consistently outperforms existing baselines, achieving substantial gains in consistency and accuracy across backbones.
\end{enumerate}

\section{Problem Formulation}

Let $\mathcal{F} = \{(s, r, o)\}$ denote the set of factual triples, where $s$ is a subject, $r$ a relation, and $o$ an object; a knowledge edit is defined as $e = (s, r, o \rightarrow o^*)$, specifying that the mapping $(s, r) \mapsto o$ should be updated to a new object $o^{*}$, and $\mathcal{E} = \{e_1, e_2, \ldots, e_m\}$ denotes the set of such edits. PPKE adds the constraint of avoiding adjustments to the model's original parameters. For MHQA under PPKE, a multi-hop question $Q$ requires reasoning over a chain of facts $C = \big[(s_1, r_1, o_1), (s_2, r_2, o_2), \ldots, (s_n, r_n, o_n)\big]$, where $s_{i+1} = o_i$ and the final answer is $o_n$; edits in $\mathcal{E}$ revise relevant triples in $C$ to form $C^* = \big[(s_1, r_1, o_1'), (s_2, r_2, o_2'), \ldots, (s_n, r_n, o_n')\big]$, with $o_i' = o_i^*$. The objective is to ensure that given $Q$, $\mathcal{F}$, and $\mathcal{E}$, the PPKE method would output an answer consistent with $C^*$.

\label{sec:assumptions}
\textbf{Assumptions} \label{sec:assumptions} We adopt the case-isolated edit, under which each edit is restricted to the current instance and does not influence other cases. \citep{zhong2023mquake} This assumption aligns with the evaluation setting and reflects practical scenarios where distinct users or domains require isolated edits that must not interfere with unrelated reasoning chains.

\section{Method}
\begin{figure}[t]
  \includegraphics[width=\columnwidth]{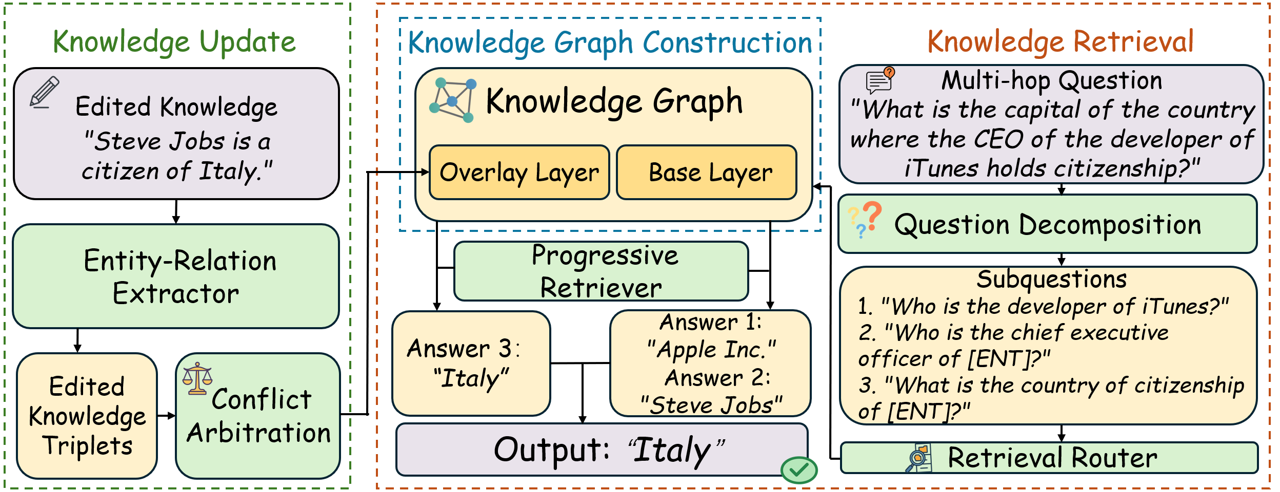}
  \caption{An overview of CAPE-KG, consists of three major components: the KG Construction module for building a pre-edit KG, the Knowledge Update module for incorporating edited knowledge, and the Knowledge Retrieval module for generating answers via retrieval.}
  \label{fig:pipeline}
  \vspace{-0.5cm}
\end{figure}

We propose the first consistency-aware PPKE framework, CAPE-KG, which addresses the fundamental challenges of three key dimensions of consistency. Fig.\ref{fig:pipeline} illustrates the overall pipeline, which consists of three major components: Knowledge Update, Knowledge Graph Construction, and Knowledge Retrieval. Each primarily ensures one form of consistency while reinforcing the others, resulting in a coherent editing framework. CAPE-KG maintains knowledge boundary consistency, where factual and edited knowledge is uncontaminated for reasoning; update consistency, where edits are incorporated in a nonconflicting and case-isolated manner; and intent consistency, where the whole retrieving process is always aligned with the purpose of the edit.

\subsection{Knowledge Graph Construction}
The construction of KG establishes the knowledge boundary consistency foundation by explicitly separating facts from edited knowledge. Knowledge boundary inconsistency arises from knowledge contamination, as factual queries mistakenly retrieve edited knowledge from unrelated cases as factual information. CAPE-KG adopts a multi-layer architecture, which includes a base layer and corresponding overlay layers. 

\textbf{Base Layer.} The base KG $\mathcal{B}$ is constructed solely from factual triples $\mathcal{F}$, dynamically derived for each batch of edit cases. By excluding all edited content, this layer establishes a consistent foundation for factual reasoning.

\textbf{Overlay Layers.} For each editing case $c$, we construct an overlay $O_c = \{(s,r,o \to o^*)\}$ containing only the edits associated with that case. Overlays are case-specific edit containers built on demand using a copy-on-write mechanism.

\textbf{Knowledge Boundary Isolation.} The separation between $B$ and $\{O_c\}$ guarantees that factual and edited knowledge remain distinct with respect to the knowledge boundary. This KG construction prevents edited knowledge from being misinterpreted as factual information, thereby maintaining knowledge boundary consistency across all reasoning processes. It provides the foundation for knowledge boundary consistency, which is further reinforced during each update and retrieval.

\label{sec:majhead}
\begin{table*}[t]
\centering
\small
\renewcommand{\arraystretch}{0.65}
\setlength{\tabcolsep}{2.75pt}

\begin{tabular}{l|l|cc|cc|cc|cc|cc}

\hline
\multicolumn{1}{c|}{\multirow{3}{*}{\textbf{Backbone}}} &
\multicolumn{1}{c|}{\multirow{3}{*}{\textbf{Method}}} &

\multicolumn{6}{c|}{\textbf{MQUAKE-CF-3K}} &
\multicolumn{4}{c}{\textbf{MQUAKE-T}} \\
\cline{3-12}
& &
\multicolumn{2}{c|}{\textbf{1 edited}} &
\multicolumn{2}{c|}{\textbf{100 edited}} &
\multicolumn{2}{c|}{\textbf{All edited}} &
\multicolumn{2}{c|}{\textbf{1 edited}} &
\multicolumn{2}{c}{\textbf{All edited}} \\
\cline{3-12}
& &
M-Acc & H-Acc & M-Acc & H-Acc & M-Acc & H-Acc & M-Acc & H-Acc & M-Acc & H-Acc \\
\hline

% ================= LLaMA 2-7B =================
\multirow{11}{*}{\shortstack{\textit{LLaMA 2-7B}}} &
FT\textsubscript{COT}*      & 22.30 & - & 2.13  & - & OOM & - & 47.32 & - & 3.75 & - \\
& FT*                        & 28.20 & 7.30 & 2.37 & 0.03 & OOM & OOM & 56.48 & 33.89 & 1.02 & 0.37 \\
& ROME\textsubscript{COT}*                             & 11.17 & - & 2.87 & - & 2.77 & - & 28.96 & - & 14.40 & - \\
& ROME*                      & 13.13 & 5.37 & 3.50 & 0.03 & 3.63 & 0.10 & 24.89 & 17.99 & 1.71 & 0.32 \\
& MEMIT\textsubscript{COT}*     & 11.83 & - & 9.23 & - & 5.57 & - & 36.88 & - & 31.58 & - \\
& MEMIT*                      & 14.97 & 6.43 & 9.40 & 2.47 & 2.30 & 0.37 & 30.89 & 23.98 & 25.21 & 20.13 \\
& MeLLo*                     & 33.57 & 9.90 & 20.00 & 10.07 & 17.33 & 9.90 & 65.78 & 55.27 & 57.69 & 44.55 \\
& PokeMQA*                      & 44.13 & 30.90 & 37.33 & 27.83 & 32.83 & 23.87 & 75.43 & 60.44 & 74.36 & 60.22 \\
& KEDKG*                           & 66.80 & 63.67 & 58.50 & 55.37 & 48.30 & 43.90 & 73.13 & 69.06 & 71.15 & 66.76 \\
& TEMPLE-MQA*               & 68.32 & 59.46 & 48.95 & 35.17 & 42.20 & 27.51 & 77.56 & 64.87 & 75.73 & 62.30 \\
& \textbf{CAPE-KG}                                     & \textbf{78.03} & \textbf{70.35} & \textbf{78.93} & \textbf{71.02} & \textbf{78.38} & \textbf{70.84} & \textbf{93.39} & \textbf{84.52} & \textbf{92.57} & \textbf{83.82} \\
\hline

% ================= Vicuna-7B =================
\multirow{5}{*}{\shortstack{\textit{Vicuna-7B}}} &
MeLLo* & 30.70 & 20.84 & 24.75 & 12.25 & 22.35 & 10.85 & 60.72 & 48.55 & 51.55 & 42.97 \\
& PokeMQA* & 45.83 & 34.80 & 38.77 & 31.23 & 31.63 & 25.30 & 74.57 & 55.19 & 73.07 & 55.09 \\
& KEDKG*     & 68.60 & 65.13 & 62.43 & 58.20 & 51.10 & 44.67 & 71.90 & 67.23 & 74.68 & 66.64 \\
& TEMPLE-MQA* & 71.61 & 62.75 & 56.65 & 44.26 & 46.60 & 37.33 & 81.77 & 69.46 & 78.29 & 68.15 \\
& \textbf{CAPE-KG}              & \textbf{81.89} & \textbf{78.74} & \textbf{81.59} & \textbf{78.85} & \textbf{80.94} & \textbf{78.49} & \textbf{96.85} & \textbf{87.25} & \textbf{94.58} & \textbf{86.23} \\
\hline

% ================= GPT-3.5 =================
\multirow{5}{*}{\shortstack{\textit{GPT-3.5}\\\textit{-turbo-instruct}}} &
MeLLo* & 57.43 & 28.80 & 40.87 & 28.13 & 35.27 & 25.30 & 88.12 & 52.84 & 74.57 & 53.53 \\
& PokeMQA* & 67.27 & 56.37 & 56.00 & 49.63 & 45.87 & 39.77 & 76.98 & 68.09 & 78.16 & 67.88 \\
& KEDKG*     & 68.00 & 65.33 & 59.50 & 56.80 & 49.10 & 43.17 & 78.75 & 76.18 & 77.19 & 73.77 \\
& TEMPLE-MQA* & 78.11 & 63.45 & 67.21 & 55.33 & 53.68 & 40.05 & 90.57 & 81.90 & 82.26 & 74.33 \\
& \textbf{CAPE-KG}              & \textbf{79.78} & \textbf{73.65} & \textbf{79.81} & \textbf{73.75} & \textbf{79.03} & \textbf{73.00} & \textbf{97.11} & \textbf{88.00} & \textbf{96.72} & \textbf{87.40} \\
\hline
\end{tabular}
\vspace{-0.2cm}
\caption{Main experimental results on the MQuAKE benchmarks \citep{zhong2023mquake} across different backbone models.
M-Acc and H-Acc denote multi-hop and holistic accuracy.
Bold numbers indicate the best performance.
Results marked with * are reported from prior work.
``OOM'' denotes out-of-memory failures due to GPU limits.}
\label{tab:mquake_results}
\vspace{-0.5cm}
\end{table*}

\subsection{Knowledge Update}
% \vspace{-0.1cm}
Knowledge update ensures knowledge edits are incorporated in a stable and case-isolated manner. Given an input edit, we adopt the entity and relation detectors introduced in KEDKG \citep{lu2025kedkg} to extract a structured triple $\{(s, r, o^*)\}$, which is then used to update the overlay of the current case. Unlike prior methods, where edits from different cases are repeatedly applied to a global graph, which leads to overwrites and unstable states, CAPE-KG enforces update consistency through a case isolation mechanism with conflict arbitration.

Updates are strictly limited to the overlay of the current case. By never propagating to the base layer or to overlays of other cases, edits remain aligned with the assumption that cases are independent. To enforce update consistency, the conflict arbitration module applies a deterministic rule: within each case, the current edit takes precedence for a given subject–relation pair; if no edit is specified, the original factual triple from the base layer is retained. This mechanism not only maintains update consistency but also reinforces the knowledge boundary separation established in construction, ensuring the boundary consistency throughout the system.

% \subsection{Ablation Studies}
\label{ssec:subhead}
\begin{table*}[t]
\centering
\footnotesize
\setlength{\tabcolsep}{7pt}
\renewcommand{\arraystretch}{0.7}

{\setlength{\tabcolsep}{2pt}
\small
\begin{tabular}{c|c|c|cc|cc|cc|cc|cc}
\hline
\multicolumn{3}{c|}{\textbf{Method}} &
\multicolumn{6}{c|}{\textbf{MQUAKE-CF-3K}} &
\multicolumn{4}{c}{\textbf{MQUAKE-T}} \\ \cline{1-13}
\multirow{2}{*}{\textbf{KG Construction}} &
\multirow{2}{*}{\textbf{Retrieval}} &
\multirow{2}{*}{\textbf{Update}} &
\multicolumn{2}{c|}{\textbf{I edited}} 
& \multicolumn{2}{c|}{\textbf{100 edited}} 
& \multicolumn{2}{c|}{\textbf{All edited}} 
& \multicolumn{2}{c|}{\textbf{I edited}} 
& \multicolumn{2}{c}{\textbf{All edited}} \\ \cline{4-13}
& & &
M-Acc & H-Acc & M-Acc & H-Acc & M-Acc & H-Acc & M-Acc & H-Acc & M-Acc & H-Acc \\ \hline

-- & \checkmark & \checkmark & 75.00 & 69.00 & 74.67 & 69.04 & 75.67 & 69.00 & 96.00 & 86.60 & 95.80  & 86.60 \\
\checkmark & -- & \checkmark & 53.33 & 43.00 & 51.33 & 46.67 & 51.93 & 46.00 & 87.33 & 81.67 & 87.67 & 82.00 \\
\checkmark  & \checkmark & --          & 77.40 & 71.00 & 74.58 & 68.14 & 74.29 & 68.10 & 96.60 & 87.10 & 94.35 & 85.60 \\ \hline
\checkmark & \checkmark & \checkmark
          & 79.78 & 73.65 & 79.81 & 73.75 & 79.03 & 73.00 & 97.11 & 88.00 & 96.72 & 87.40 \\ \hline
\end{tabular}}
\vspace{-0.2cm}
\caption{Ablation study of CAPE-KG on MQuAKE-CF-3K and MQuAKE-T \citep{zhong2023mquake}. Each component incrementally contributes to improved consistency and accuracy, with the full framework achieving the best results.}
\label{tab:ablation_results}
\vspace{-0.4cm}
\end{table*}
\subsection{Knowledge Retrieval}
\vspace{-0.1cm}
In the retrieval stage, an input query $Q$ is decomposed into a sequence of sub-questions $\{q_1, q_2, \ldots, q_n\}$ using a similarity-based dynamic few-shot strategy \citep{izacard2022atlas}. Existing methods often fall back to unedited factual knowledge through LLM responses when retrieval fails, which can produce answers that contradict the intended edit. To address this, our framework designs a retrieval process that explicitly routes queries between the base and overlay layers, applies progressive strategies when uncertainty arises, and suppresses irrelevant edits. 

\textbf{Edit-Aware Retrieval Routing.} We define the edit impact surface for case $c$ as $S_{\text{edit}}^c = \{s \mid (s, r, o^*) \in E^c\}$ and $P_{\text{edit}}^c = \{r \mid (s, r, o^*) \in E^c\}$, where $E^c$ represents the edit set for case $c$. For each sub-question $q_i$, we determine the target knowledge layer:
{\setlength{\abovedisplayskip}{6pt}
 \setlength{\belowdisplayskip}{6pt}
\begin{equation}
\small
L(q_i)=
\begin{cases}
\mathrm{Overlay}^c\!, &
\operatorname{subject}(q_i)\in S_{\text{edit}}^c \\[-2pt]
& \quad \lor\ \operatorname{relation}(q_i)\in P_{\text{edit}}^c \\
\mathrm{Base}\!, & \text{otherwise}
\end{cases}
\end{equation}
}

This routing ensures that factual queries access the Base layer while edit-sensitive queries access the current case's Overlay layer.

\textbf{Progressive Retrieval.} When answering a sub-question, the system may encounter uncertainty regarding entity or relation matching. To address this, we introduce a progressive retrieval strategy with three stages.

High Confidence Stage: For entity candidates $K = \{k_1, \ldots, k_n\}$ with scores $g_{\phi}(q_{i}, k_{j}) \in [0,1]$ generated by the detector, we first select those above a threshold $\tau$, $K' = { k_{j} \in K \mid g_{\phi}(q_{i}, k_{j}) \geq \tau }$. We then filter out low outliers using mean $\mu$ and deviation $\sigma$, keeping $K'' = { k_{j} \in K' \mid g_{\phi}(q_{i}, k_{j}) \geq \mu - \lambda \sigma }$ where $\lambda > 0$ is adaptive. Edit irrelevance suppression is performed during entity scoring, where confidence scores of edited entities not explicitly referenced in the sub-question are down-weighted to prevent unintended influence. The remaining candidates are sorted and used sequentially for retrieval, ensuring stable results when multiple entities have similar scores.

Low Confidence Stage: When no candidates exceed the threshold, i.e., $K'=\emptyset$, we invoke a language model to select the most plausible entity from the original candidate pool $K$.

Failure Stage: If no valid entity is retrieved, a relation detector checks whether the subquestion $q_i$ falls within the scope of an edit. When relevant, the LLM is prompted with the full edited triple $(s, r, o^*)$ injected into the context; otherwise, it generates an answer from $q_i$ alone.

Together, intent consistency is ensured, where the reasoning process is always aligned with the purpose of the edit. Illustrative examples can be found in Appendix~\ref{sec:case}.

\section{Experiment}
\subsection{Experiment Setting}
\textbf{Baselines.} We compare CAPE-KG against parameter-updating editors (FT \citep{zhu2020modifying}, ROME \citep{meng2022locating}, MEMIT \citep{meng2023mass}) and parameter-preserving methods including MeLLo \citep{zhong2023mquake}, PokeMQA \citep{gu2024pokemqa}, KEDKG \citep{lu2025kedkg}, and TEMPLE-MQA \cite{cheng2024temporal}. Additional details are provided in Appendix~\ref{app:baselines}.

\textbf{Setup.} Experiments are conducted on benchmarks MQuAKE-CF-3K and MQuAKE-T \citep{zhong2023mquake}; dataset details are in Appendix~\ref{sec:dataset}.  Retrieval parameters $\tau$, $\lambda$ are selected via grid search in Appendix~\ref{app:hyperparameters}. Other details are in Appendix~\ref{app:experiment}.

\textbf{Evaluation Metrics.} The framework is evaluated by  Match Accuracy (M-Acc), which measures the correctness of the final predicted answer against the benchmark ground truth, and Hop Accuracy (H-Acc), which measures the accuracy of intermediate hops against annotated ground-truth chains. 

\subsection{Results}

\textbf{Main Results.} Table~\ref{tab:mquake_results} shows that CAPE-KG consistently outperforms all baselines across all backbones and datasets. On average, we achieve an absolute improvement of 18.32\% in M-Acc and 19.16\% in H-Acc on MQuAKE-CF-3K, an absolute improvement of 14.17\% in M-Acc and 10.14\% in H-Acc on MQuAKE-T. These substantial gains demonstrate the effectiveness of enforcing multiperspective consistency in PPKE for MHQA. The improvements hold across LLaMA 2 \citep{touvron2023llama2}, Vicuna \citep{zheng2024judge}, and GPT-3.5 backbones, indicating that the performance gains in accuracy are model-independent and arise from this consistency-aware architecture.

\textbf{Consistency Across Batch Edits.} We observed accuracy drops in both M-Acc and H-Acc on baseline as edit batch size increases, due to overwrites and cross-case contamination. In addition to higher accuracy, CAPE-KG's multi-layer design with isolated overlays preserves consistency under larger batch sizes, leading to smaller fluctuations. The absolute gap between single-edit and all-edit settings across two datasets averages only 0.91\% for M-Acc and 0.62\% for H-Acc, further highlighting the improvement in consistency of our approach.

\subsection{Ablation Studies}
Table~\ref{tab:ablation_results} shows the ablation study of CAPE-KG by disenabling three components: KG Construction, Knowledge Update, and Knowledge Retrieval. Performance improves steadily as more modules are introduced. Among them, the Retrieval module contributes the most significant gains, highlighting the importance of enforcing intent consistency. In contrast, KG Construction and Knowledge Update have relatively minor impacts, as their effectiveness depends on the prevalence of conflicts in the dataset. Nevertheless, the combination of all three modules results in the highest accuracy, demonstrating that multiperspective consistency is best achieved when the full framework is employed, which makes the most effective PPKE on MHQA.

\section{Conclusion}
We propose CAPE-KG, which advances PPKE for MHKE by enforcing knowledge boundary consistency, update consistency, and intent consistency through components of the framework, including KG construction, knowledge update, and knowledge retrieval. In future work, we expect to extend CAPE-KG to continual knowledge editing, ensuring consistency as edits evolve over time.

\section*{Limitations}
The progressive retrieval strategy introduces additional inference latency compared to single-pass retrieval, particularly under deeper reasoning chains. This overhead reflects a deliberate trade-off for improved consistency, and a detailed latency analysis is provided in Appendix~\ref{sec:latency}.

As mentioned in Section~\ref{sec:assumptions}, our method relies on a case-isolated edit assumption, which is suitable for practical scenarios in which different users require distinct edits that must remain isolated to prevent cross-user contamination. However, this assumption does not hold in settings where knowledge updates are globally shared, which are beyond the scope of CAPE-KG.

% Bibliography entries for the entire Anthology, followed by custom entries
%\bibliography{anthology,custom}
% Custom bibliography entries only
\bibliography{main}

\appendix
\vspace{0.1cm}
\section{Dataset}
\label{sec:dataset}

Table~\ref{tab:dataset_statistics} summarizes the statistics of the two benchmark datasets used for evaluation, including \textsc{MQuAKE-CF-3K} and \textsc{MQuAKE-T} \citep{zhong2023mquake}. \textsc{MQuAKE-CF-3K} is constructed from counterfactual knowledge edits to evaluate whether models can propagate hypothetical factual changes through multi-hop reasoning, containing 3,000 questions spanning 2–4 hops across multiple edited facts. \textsc{MQuAKE-T} consists of 1,868 instances derived from real-world temporal knowledge updates, assessing reasoning under time-sensitive factual changes.

\begin{table}[H]
    \centering
    \small
    \setlength{\tabcolsep}{4pt}
    \begin{tabular}{lccccc}
        \toprule
        Dataset & \#Edits & 2-hop & 3-hop & 4-hop & Total \\
        \midrule
        \multirow{5}{*}{MQuAKE-CF-3K} 
        & 1   & 513  & 356  & 224  & 1093 \\
        & 2   & 487  & 334  & 246  & 1067 \\
        & 3   & --   & 310  & 262  & 572  \\
        & 4   & --   & --   & 268  & 268  \\
        & All & 1000 & 1000 & 1000 & 3000 \\
        \midrule
        MQuAKE-T & 1 (All) & 1421 & 445 & 2 & 1868 \\
        \bottomrule
    \end{tabular}
    \caption{Statistics of the multi-hop question answering datasets MQuAKE used in our experiments.}
    \label{tab:dataset_statistics}
\end{table}

\section{Baselines}
\label{app:baselines}
We compare \textsc{CAPE-KG} with two categories of baseline methods: 
(1) parameter-updating knowledge editing methods and 
(2) parameter-preserving knowledge editing methods. 
All baselines are evaluated under the same experimental settings for fair comparison.

\subsection{Parameter-Updating Editing Methods}

\paragraph{FT} 
\citep{zhu2020modifying} fine-tunes model parameters on edited facts via gradient-based optimization to incorporate new knowledge.

\paragraph{ROME} 
\citep{meng2022locating} identifies the model parameters associated with a given factual relation and applies a low-rank update to the feed-forward network of a selected transformer layer to encode the edited knowledge.

\paragraph{MEMIT} 
\citep{meng2023mass} extends ROME by applying coordinated low-rank updates across multiple feed-forward layers of the transformer to encode edited knowledge.

\subsection{Parameter-Preserving Editing Methods}

\paragraph{MeLLo} 
\citep{zhong2023mquake} adopts a plan-and-solve framework that decomposes multi-hop questions into sub-questions and retrieves edited facts to guide intermediate reasoning without modifying model parameters.

\paragraph{PokeMQA} 
\citep{gu2024pokemqa} extends MeLLo by introducing a detached scope detector that determines whether a sub-question is affected by edited knowledge and modulates the retrieval process accordingly.

\paragraph{KEDKG} 
\citep{lu2025kedkg} encodes edited knowledge in an external knowledge graph and performs graph-based retrieval and reasoning at inference time without modifying model parameters.

\paragraph{TEMPLE-MQA} 
\citep{cheng2024temporal} models temporal dependencies in multi-hop question answering and incorporates edited knowledge through structured retrieval while preserving model parameters.

\section{Illustrative Analysis of CAPE-KG Consistency}
\label{sec:case}

\subsection{Knowledge Boundary Consistency}
\label{app:boundary}
\begin{figure}[t!]
    \centering
    \includegraphics[width=1\linewidth]{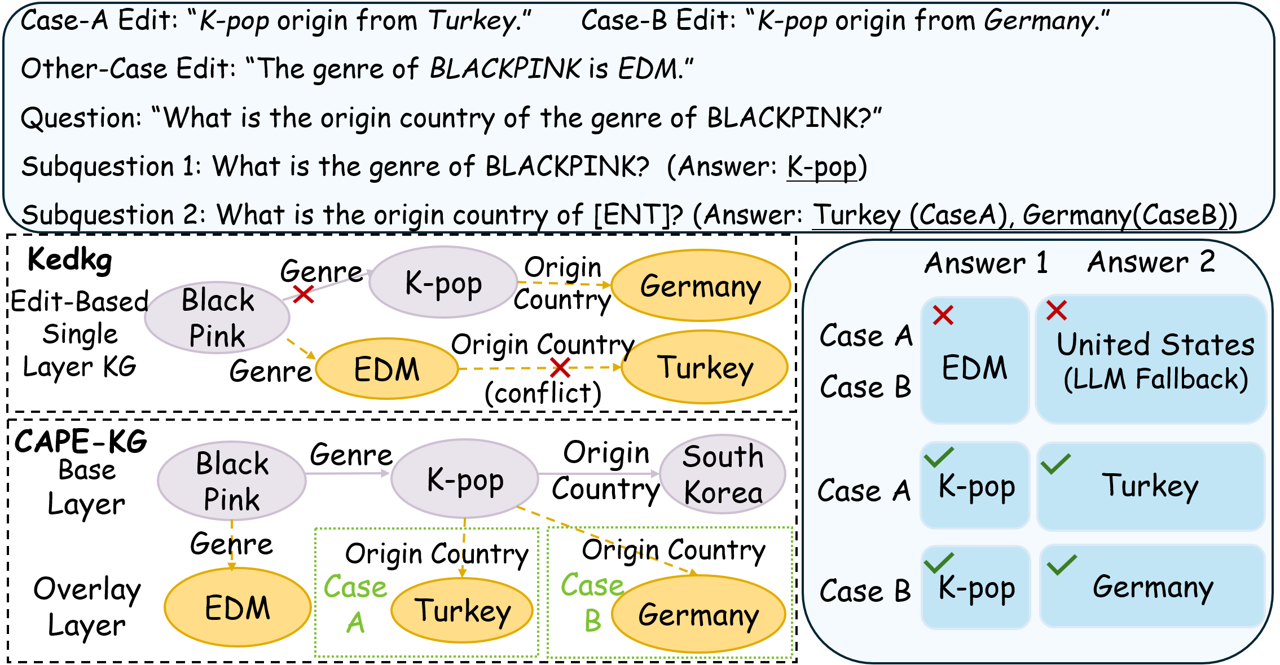}
    \caption{Case Comparison of Update and KG Construction. Unedited knowledge is labeled in purple. Edited entities are in yellow, with dotted lines indicating the edited relation.}
    \label{fig: update_case}
\end{figure}

CAPE-KG has effectively improved the knowledge boundary consistency by avoiding possible contamination. In the case shown in Fig.~\ref{fig: update_case}, prior work has the potential inconsistency of mixing unrelated edit from other cases, (Black Pink, genre, EDM), into reasoning as fact. EDM is retrieved instead of K-pop, which breaks the reasoning chain, while CAPE-KG maintains knowledge boundary consistency by using this multi-layer design.

\subsection{Update Consistency}
\label{app:update}
CAPE-KG prevents the possible inconsistency caused by overwriting conflicting edits in a global KG. In the case shown in Fig.~\ref{fig: update_case}, two edits (K-pop,
origin country, Turkey), (K-pop, origin country, Germany), so one remains while the other is discarded. 
\textsc{CAPE-KG} stores each edit in a case-specific overlay, preserving both without interference. This design keeps per-case reasoning aligned with the case-isolated edit assumption and ensures update consistency.

\subsection{Intent Consistency}
\label{app:intent}
\begin{figure}[t!]
    \centering
    \includegraphics[width=1\linewidth]{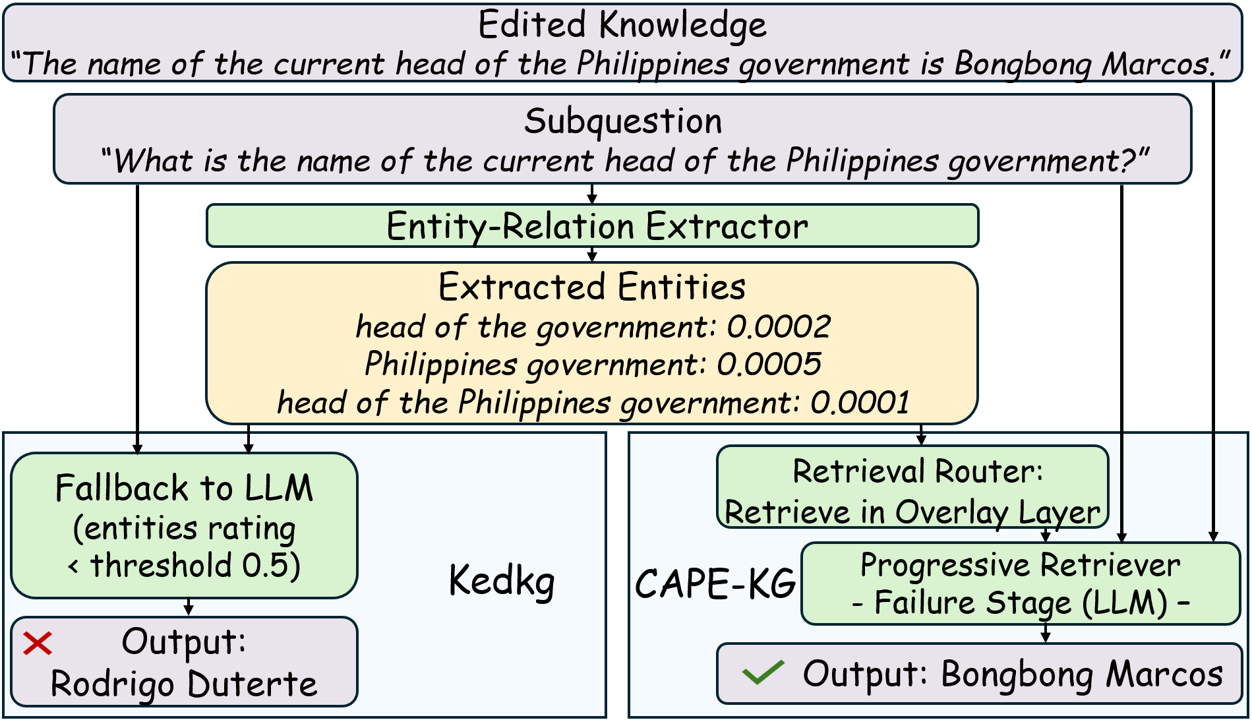}
    \caption{Case Comparison of Intent Consistency in Retrieval.}
    \vspace{-0.3cm}
    \label{fig:retrieval_case}
\end{figure}

As shown in Fig.~\ref{fig:retrieval_case}, when a subquestion is directly related to an edited fact, prior work falls back to LLM answering directly under low-confidence entity extraction. This fallback would query unedited factual knowledge, which is inconsistent with the intent of editing. In contrast, CAPE-KG routes edit-related queries to the overlay layer via the retrieval router, ensuring that edits are explicitly applied in reasoning. Moreover, at the failure stage of progressive retrieval, the LLM is also queried, but with the edit injected into the query context. This guarantees that even when a component of CAPE-KG is uncertain, the reasoning process remains aligned with the intended update, in this way preserving intent consistency.

\section{ Prompts}
\label{sec:prompts}

\begin{table}[t]
\centering
\small
\setlength{\tabcolsep}{6pt}
\begin{tabular}{|>{\raggedright\arraybackslash}m{0.95\columnwidth}|}
\hline
\rule{0pt}{0.1cm}

You are an expert at decomposing multi-hop questions into sequential sub-questions. \\
Each sub-question should be answerable independently, and the answer from one sub-question \\
is passed to the next sub-question using the [ENT] placeholder. \\
\\
Rules: \\
1. Break down complex questions into simple, sequential sub-questions \\
2. Use [ENT] to pass entities between sub-questions \\
3. Each sub-question should focus on one specific fact \\
4. Maintain logical flow from general to specific \\
5. Avoid redundant or irrelevant questions \\
\\
Question:\\<<<<DEMO QUESTION 1>>>> \\
Sub-question: \\
<<<<DEMO SUB-QUESTION 1>>>> \\
\\
Question: \\<<<<DEMO QUESTION 2>>>> \\
Sub-question: \\
<<<<Demo SUB-QUESTION 2>>>> \\
\\
\textnormal{[Additional dynamically retrieved demonstrations omitted]} \\
\\
Decompose the following question:\\
Question:\\ <<<<QUESTION>>>> \\
Sub-question: \\
\\
\hline
\end{tabular}
\caption{Prompt template for multi-hop question decomposition.}
\label{tab:decomposition_prompt}
\end{table}

\begin{table}[t]
\centering
\small
\setlength{\tabcolsep}{6pt}
\begin{tabular}{|>{\raggedright\arraybackslash}m{0.95\columnwidth}|}
\hline
\rule{0pt}{0.2cm} 

You are an expert at answering factual questions with precise and concise entity names. \\
Your answers should be specific entities that can be used in subsequent questions. \\
\\
Rules: \\
1. Answer with the most specific and complete entity name \\
2. Use proper capitalization and full names \\
3. Avoid articles (a, an, the) unless part of the name \\
4. Be consistent with entity naming \\
5. If multiple answers exist, choose the most relevant one \\
6. If uncertain, provide the most likely answer \\
\\
Examples: \\
Question: Which country was jazz created in? \\
Answer: United States of America \\
\\
Question: What is the capital of France? \\
Answer: Paris \\
\\
Question: Who wrote the play ``Hamlet''? \\
Answer: William Shakespeare \\
\\
\textnormal{[Additional in-context demonstrations omitted]} \\
\\
Answer the following question: \\
Question: <<<<SUB-QUESTION>>>> \\
Answer: \\
\\
\hline
\end{tabular}
\caption{Prompt template for factual sub-question answering.}
\label{tab:answer_prompt}
\end{table}

CAPE-KG has three prompt templates: one for multi-hop question decomposition (Table~\ref{tab:decomposition_prompt}) and two for sub-question answering (Table~\ref{tab:answer_prompt}, Table~\ref{tab:edit_prompt}).

For question decomposition, we adopt a retrieval-based few-shot prompting strategy in Table~\ref{tab:decomposition_prompt} using a K-nearest-neighbor (KNN) \cite{cover1967nearest} search over a fixed reference pool.  Few-shot question decomposition demos are drawn from MQuAKE-CF \citep{zhong2023mquake} after removing overlaps with evaluation data.
For each input question, the top-k most similar examples (k=6 in our experiment) are retrieved based on semantic similarity computed using frozen sentence embeddings. The retrieved demonstrations are attached to a fixed instruction template to form the final decomposition prompt, providing contextually relevant guidance without updating model parameters.

For sub-question answering, when a sub-question does not fall within the scope of any edit, and no valid entity can be retrieved from the knowledge graph, the language model is prompted to directly answer the sub-question based on its internal factual knowledge, without any injected edits. In this case, we use an instruction-based prompt in Table~\ref{tab:answer_prompt} with fixed few-shot demonstrations to encourage concise, entity-level outputs that can be consumed by subsequent reasoning steps. This fallback mechanism enables the model to provide missing factual information when explicit knowledge retrieval fails, while avoiding the introduction of unintended edits.

\begin{table}[t]
\centering
\small
\setlength{\tabcolsep}{6pt}
\begin{tabular}{|>{\raggedright\arraybackslash}m{0.95\columnwidth}|}
\hline
\rule{0pt}{0.6cm} 
New Fact: <<<<EDITED KNOWLEDGE>>>> \\
\\
Question: <<<<SUB-QUESTION>>>> \\
Answer: \\
\\
\hline
\end{tabular}
\caption{Prompt template for edit-relevant sub-question answering.}
\vspace{-0.3cm}
\label{tab:edit_prompt}
\end{table}

For sub-question answering, when a sub-question is edit-relevant and knowledge graph retrieval fails, we adopt an in-context learning formulation following prior work ~\citep{zheng2023ike} to inject edited knowledge into the language model input. Specifically, the edited triple $(s, r, o^*)$ is filled into a fixed instruction template, as shown in Table~\ref{tab:edit_prompt}, and explicitly injected into the input context. Such edit-aware prompting ensures that fallback answering remains aligned with the intended update even when explicit retrieval fails, while preserving strict case isolation and parameter preservation.

\section{Details about Experiments}
\label{app:experiment}
All experiments in this paper are conducted on English-language benchmarks MQuAKE  \citep{zhong2023mquake}, and all language models and prompts operate exclusively in English. Experiments are run on a single machine equipped with an NVIDIA RTX 5080 GPU. LLaMA-2-7B \citep{touvron2023llama2} and Vicuna-7B \citep{zheng2024judge} with 7 billion parameters each, and GPT-3.5-turbo-instruct with an undisclosed parameter size, are used as backbone models. For each experimental setting, we conduct a single full run over the entire benchmark and report accuracy averaged across all instances. Accordingly, all reported M-Acc and H-Acc results correspond to single-run benchmark-level averages, without additional variance estimates.

\begin{table*}[!t]
\centering
\small
\setlength{\tabcolsep}{3pt}
\renewcommand{\arraystretch}{0.95}
\begin{tabular}{
    >{\centering\arraybackslash}m{5.0cm}
    | c | c | c | c | c | c | c
}
\hline
& \multirow{2}{*}{\textbf{Single-Pass}}
& \multirow{2}{*}{\textbf{Progressive}}
& \multicolumn{3}{c|}{\textbf{MQuAKE-CF-3K}}
& \multicolumn{2}{c}{\textbf{MQuAKE-T}} \\
\cline{4-8}
&  &
& \textbf{1 edited} & \textbf{100 edited} & \textbf{All edited}
& \textbf{1 edited} & \textbf{All edited} \\
\hline
\multirow{2}{*}{\textbf{Avg. Latency per Multi-Hop QA (ms)}}
& $\checkmark$ &
& 5177.98 & 4737.47 & 5061.52 & 4776.70 & 4588.83 \\
\cline{2-8}
&  & $\checkmark$
& 6203.85 & 7174.07 & 6041.60 & 6342.65 & 6071.88 \\
\hline
\multirow{2}{*}{\textbf{Avg. Latency per Sub-Question (ms)}}
& $\checkmark$ &
& 136.98 & 133.21 & 172.68 & 146.16 & 139.93 \\
\cline{2-8}
&  & $\checkmark$
& 201.63 & 226.34 & 193.77 & 238.63 & 227.58 \\
\hline
\end{tabular}
\caption{Latency comparison between single-pass and progressive retrieval across edit batch sizes on MQuAKE benchmarks \citep{zhong2023mquake}.}
\label{tab:latency_analysis}
\end{table*}

\section{Progressive Retrieval Latency}
\label{sec:latency}

 While progressive retrieval enhances answer accuracy by dynamically updating the knowledge graph during multi-hop reasoning, it introduces additional computational overhead. To quantify this computational overhead, we conduct a latency analysis comparing our full method with progressive retrieval against a variant without progressive updates. We report a detailed latency breakdown in Table~\ref{tab:latency_analysis}.
As discussed in the main paper, progressive retrieval gradually retrieves and integrates evidence across multiple reasoning steps, which is essential for enforcing consistency under multi-hop reasoning and multiple edits.
Our ablation studies confirm its effectiveness for the target task.

To analyze this overhead, we evaluate inference latency under both progressive retrieval and single-pass retrieval across different edit batch sizes on \textsc{MQuAKE-CF-3K} and \textsc{MQuAKE-T}  \citep{zhong2023mquake}. We report two latency metrics: (1) the average end-to-end latency per multi-hop question, and (2) the average latency per sub-question. All experiments are conducted using GPT-3.5-turbo-instruct as the backbone language model, under the same datasets, edit configurations, and experimental settings as the main experiments.

As shown in Table~\ref{tab:latency_analysis}, progressive retrieval results in higher inference latency than single-pass retrieval.
This latency is expected, as progressive retrieval performs complex retrieval and reasoning steps to enforce consistency under multi-hop edits.
While this introduces additional computational cost, inference efficiency is not the primary objective of this work.
Importantly, our ablation studies demonstrate that progressive retrieval is essential for achieving reliable and consistent multi-hop knowledge editing, and we therefore view the observed latency as a reasonable and deliberate trade-off for improved effectiveness.

% preamble:
% \usepackage{array}
% \usepackage{multirow}

% preamble:
% \usepackage{array}
% \usepackage{multirow}

\section{Hyperparameter Settings}
\label{app:hyperparameters}

We conduct a hyperparameter analysis for $\tau$ and $\lambda$, with values selected via grid search using GPT-3.5-turbo-instruct on MQuAKE-CF-3K \citep{zhong2023mquake} under the single-edit setting.
Table~\ref{tab:lambda_sensitivity} reports the results for $\lambda$, while Table~\ref{tab:tau_sensitivity} summarizes the evaluation of $\tau$.
Based on this analysis, we set $\tau = 0.4$ and $\lambda = 1.0$ in all main experiments.

\vspace{-0.2cm}
\begin{table}[H]
\centering
\small
\setlength{\tabcolsep}{6pt}
\renewcommand{\arraystretch}{0.7}
\begin{tabular}{c c}
\toprule
\textbf{$\lambda$} & \textbf{M-Acc} (\%) \\
\midrule
0.5 & 77.83 \\
\textbf{1.0} & \textbf{79.78} \\
1.5 & 79.46 \\
2.0 & 79.28 \\
2.5 & 78.52 \\
\bottomrule
\end{tabular}
\vspace{-0.2cm}
\caption{Hyperparameter analysis for $\lambda$.}
\label{tab:lambda_sensitivity}
\end{table}

\vspace{-0.2cm}
\begin{table}[H]
\centering
\small
\setlength{\tabcolsep}{6pt}
\renewcommand{\arraystretch}{0.7}
\begin{tabular}{c c}
\toprule
\textbf{$\tau$} & \textbf{M-Acc (\%)} \\
\midrule
0.1 & 77.58 \\
0.2 & 78.36 \\
0.3 & 79.52 \\
\textbf{0.4} & \textbf{79.78} \\
0.5 & 79.05 \\
0.6 & 79.30 \\
0.7 & 78.93 \\
0.8 & 78.38 \\
0.9 & 77.75 \\
\bottomrule
\end{tabular}
\vspace{-0.2cm}
\caption{Hyperparameter analysis for $\tau$.}
\label{tab:tau_sensitivity}
\end{table}

\section{Scientific Artifacts Used}
\label{app:artifact}
\paragraph{\textbf{MQuAKE.}}
We use the MQuAKE benchmark \citep{zhong2023mquake}, includes MQuAKE-CF-3K and MQuAKE-T, for evaluating multi-hop question answering under knowledge editing. According to the official release, the MQuAKE datasets and accompanying code are released under the MIT License. Their use in this work complies with its terms for academic research purposes. The benchmark is constructed from structured knowledge sources rather than user-generated text. Based on the dataset documentation and its established use in prior work, we verified that it does not contain personal data or identifiable individual information, and therefore no anonymization or offensive-content filtering was required.

\paragraph{\textbf{Pretrained Language Models.}}
We use pretrained large language models as backbones in our experiments, including LLaMA-2-7B \citep{touvron2023llama2}, Vicuna-7B \citep{zheng2024judge}, and GPT-3.5-turbo-instruct. LLaMA 2 and Vicuna are released under the LLaMA 2 Community License, and GPT-3.5-turbo-instruct is accessed via API in compliance with OpenAI’s terms of use. All models are used without modifying their original parameters and are used for academic research purpose, which is consistent with their intended use. As pretrained models are used without access to or modification of their training data, no additional checks or anonymization related to personal or offensive content are applicable at the model level.

\end{document}